\documentclass[
10pt, 
a4paper, 
oneside, 
headinclude,footinclude, 
BCOR=5mm, 
]{article} 

\usepackage[
nochapters, 
beramono, 
eulermath,
pdfspacing, 
dottedtoc 
]{classicthesis} 

\usepackage{arsclassica} 

\usepackage[T1]{fontenc} 

\usepackage{graphicx} 
\graphicspath{{Figures/}} 

\usepackage{enumitem} 

\usepackage{lipsum} 

\usepackage{subfig} 

\usepackage{amsmath,amssymb,amsthm} 

\usepackage{varioref} 

\usepackage[citestyle=authoryear,natbib=true,backend=bibtex, sorting=ynt]{biblatex}

\bibliography{references.bib}

\theoremstyle{definition} 

\theoremstyle{plain} 

\theoremstyle{remark} 

\hypersetup{
colorlinks=true, breaklinks=true, bookmarks=true, bookmarksnumbered,
urlcolor=webbrown, linkcolor=RoyalBlue, citecolor=RoyalBlue, 
pdftitle={}, 
pdfauthor={\textcopyright}, 
pdfsubject={}, 
pdfkeywords={}, 
pdfcreator={pdfLaTeX}, 
pdfproducer={LaTeX with hyperref and ClassicThesis} 
} 

\title{Text Classification of Manifestos and COVID-19 Press Briefings using BERT and Convolutional Neural Networks}

\author{\spacedlowsmallcaps{Kakia Chatsiou*}} 

\date{\small\today} 

\begin{document}


\renewcommand{\sectionmark}[1]{\markright{\spacedlowsmallcaps{#1}}} 
\lehead{\mbox{\llap{\small\thepage\kern1em\color{halfgray} \vline}\color{halfgray}\hspace{0.5em}\rightmark\hfil}} 

\pagestyle{scrheadings} 


\maketitle 

\setcounter{tocdepth}{2} 


\let\thefootnote\relax\footnotetext{* \textit{ESRC Business and Local Government Data Research Centre, School of Computer Science and Electronic Engineering, University of Essex, Colchester, United Kingdom. Manuscript of \today. Comments welcome to achats@essex.ac.uk. This work is still in progress, the usual disclaimers about errors, omissions etc apply.}}


\begin{abstract}
We build a sentence-level political discourse classifier using existing human expert annotated corpora of political manifestos from the Manifestos Project \parencite{volkens_manifesto_2020} and applying them to a corpus of COVID-19 Press Briefings \parencite{chatsiou_covid-19_2020}. We use manually annotated political manifestos as training data to train a local topic Convolutional Neural Network (CNN) classifier; then apply it to the COVID-19 Press Briefings Corpus to automatically classify sentences in the test corpus. We report on a series of experiments with CNN trained on top of pre-trained embeddings for sentence-level classification tasks. We show that CNN combined with transformers like BERT outperforms CNN combined with other embeddings (Word2Vec, Glove, ELMo) and that it is possible to use a pre-trained classifier to conduct automatic classification on different political texts without additional training.
\end{abstract}



\section{Introduction}
A substantial share of citizen involvement in politics arises through written discourse especially in the digital space. Through advanced, novel communication strategies, the public can play their part in constructing a political agenda, which has led politicians to increasingly use social media and other types of digital broadcasting to communicate (compared to mainstream press and traditional print media). This is especially pertinent with crisis communication discourse and the recent COVID-19 pandemic has created a great opportunity to study how similar topics get communicated in different countries and the narrative choices made by government and public health officials at different levels of governance (international, national, regional). To aid fellow scholars with the systematic  study of such a large and dynamic set of unstructured data, we set out to employ a text categorization classifier trained on similar domains (like existing manually annotated sentences from political manifestos) and use it to classify press briefings about the pandemic in a more effective and scalable way. 

The main attraction behind using manually coded political manifestos \parencite{volkens_manifesto_2020} as training data is that the political science expert community have been manually collecting and annotating in a systematic way political parties’ manifestos for years (since the 1960s) around the world in order to apply content analysis methods and to advance political science. They have subsequently been used as training data in semi-supervised domain-specific  classification tasks with good results \parencite{zirn_classifying_2016, nanni_topfish_2016, glavas_cross-lingual_2017, bilbao-jayo_automatic_2018, bilbao-jayo_political_2018}.

In this paper, we build variations of a CNN sentence-level political discourse classifier using existing annotated corpora of political manifestos from the Manifestos Project \parencite{volkens_manifesto_2020}. We test different CNN and word embedding architectures on the already annotated (english language) sentences of the Manifestos Project Corpus. We then apply them to a corpus of COVID-19 Press Briefings \parencite{chatsiou_covid-19_2020}, a subset of which was manually annotated by political scholars for the purposes of this work. 

The article is organised as follows: we first offer a brief overview of previous related work on the use of human expert annotated political manifestos for discourse classification. 
We then describe our framework including the training data used, data pre-processing performed and used architecture. We report on a series of experiments with CNN trained on top of pre-trained word vectors for sentence-level classification tasks. 
We conclude with evaluation of the BERT+CNN architecture against other combinations (Word2Vec+CNN, GloVe+CNN, ELMo+CNN) for both corpora.  Experimental results show  that a CNN classifier combined with transformers like BERT outperforms CNN combined with other non-context sensitive embeddings (Word2Vec, Glove, ELMo).

\section{Related Work}

The use of NLP methods to analyse political texts is a well-established field within Political Science and Computational Social science more generally \parencite{lazer_life_2009, grimmer_text_2013, benoit_treating_2009}.

Researchers have used NLP methods to acccomplish various classification tasks, such as \textit{political positioning} on a left to right continuum \parencite{slapin_scaling_2008, glavas_cross-lingual_2017}, \textit{identification of political ideology differences from text} \parencite{sim_measuring_2013, menini_agreement_2016}, \textit{detection of political events} \parencite{nanni_building_2017}, or \textit{detection of opinion and sentiment} \parencite{young_affective_2012}. 

\subsection{Topic Classification of political discourse}

A substantial body of recent work has focused on topic classification in political texts \parencite{lauscher_entities_2016, baturo_understanding_2017} some using supervised models \parencite{purpura_automated_2006, stewart_use_2009, benoit_crowd-sourced_2016, glavas_cross-lingual_2017}, others using unsupervised models such as latent semantic analysis \parencite{hofmann_probabilistic_1999} and latent Dirichlet allocations (LDA) \parencite{blei_latent_2003} or structural topic modelling \parencite{lindstedt_structural_2019, jacobs_topic_2019}

Topic classification of domain-specific types of political text, such as \textit{political manifestos} and their use as training data for unsupervised methods is receiving increased attention.  

\citet{zirn_classifying_2016}  independently trained three sentence-level classifiers - one for detecting the topic and two for detecting topic-shifts - and then combined their predictions in a global optimisation setting using a Markov Logic Network. Their experimental results show that the proposed global model achieves high classification performance and significantly outperforms the local sentence-level topic classifier.

\citet{glavas_cross-lingual_2017} propose an approach for cross-lingual topical coding of sentences from electoral manifestos using as training data, manually coded manifestos with a total of 77500 sentences in four languages (English, French, German and Italian) (and CNNs with word embeddings) and inducing a joint multilingual embedding space. They report achieving better results than monolingual classifiers in English, French and Italian but worse results with their multilingual classifier than a monolingual classifier in German.

More recently, \citet{bilbao-jayo_automatic_2018} build a sentence classifier using multi-scale convolutional neural networks trained in seven different languages trained with sentences extracted from annotated parties’ election manifestos. They use the full range of the domains defined by the manifestos project and they prove that enhancing the multi-scale convolutional neural networks with context data improves their classification. For a detailed discussion of different deep learning text classification-based models for text classification and their technical contributions, similarities, and strengths  \parencite[see]{chatsiou_deep_2020, minaee_deep_2020}.

\paragraph{Domain transfer of political manifestos classification to other political texts}

Using annotated political manifestos as the training dataset for classifying other types of political texts is gaining traction in the literature, especially with the boost in performance of deep learning methods for text. 

\citet{nanni_topfish_2016} used expert annotated political manifestos in English and speeches to train a local supervised topic classifier (SVM with a bag of words approach) that combines lexical with semantic textual similarity features at a sentence-level. A sub-part of the training set was annotated manually by human experts, and the rest was labelled automatically with the global optimisation step performed via a Markov Logic network presented in \citet{zirn_classifying_2016}. The advantage of such a domain transfer approach is that no manual topic annotation on the rest of the corpus is needed. They then classify the speeches from the 2008, 2012 and 2016 US presidential campaign into the 7 domains defined by the Manifestos Project, without the need for additional topic annotation.

\citet{bilbao-jayo_political_2018} used annotated political manifestos in Spanish and the Regional Manifestos Project taxonomy \citet{alonso_measuring_2013}, to train a neural network sentence-level classifier (CNN) with Word2Vec word embeddings, also taking account the context of the phrase (like what was previously said and the political affiliation of the transmitter). They used  this to analyse social media (twitter) data of the main Spanish political parties during 2015 and 2016 Spanish general elections without the need for additional manual coding of the twitter data.

This paper builds on this area of research presenting a comparison of a CNN classifier trained on the manifestos project annotations for English, but comparing more context-free (Word2Vec, Glove, ELMo) to context-sensitive (BERT) word embeddings. We then apply this to a corpus of daily press-briefings on the COVID-19 status by government and public health authorities.

\subsection{Datasets}

\paragraph{Manifestos Project Corpus}

\begin{figure*}[t]
\centering
\includegraphics[width=1 \textwidth]{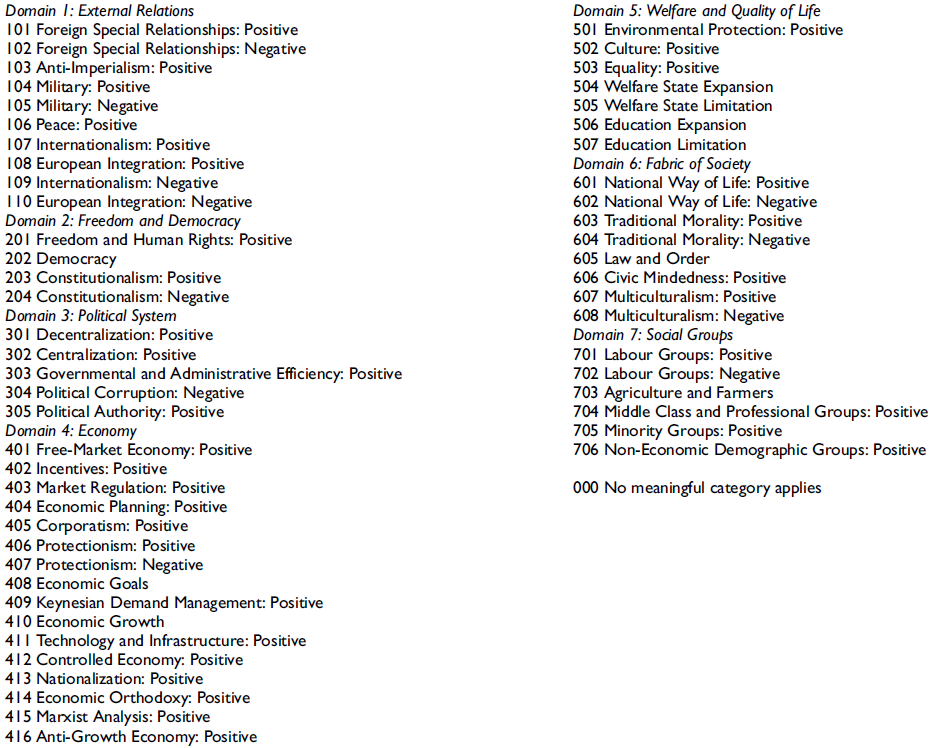} 
\caption{Manifestos Project annotation domains and subdomains used by human expert  annotators \parencite{volkens_manifesto_2020-1}, taken from \citet{bilbao-jayo_automatic_2018}}
\label{fig1-manifestos-domains}
\end{figure*}

The main attraction behind using manually coded political manifestos \parencite{volkens_manifesto_2020} as training data is that the political science community has been manually collecting and annotating in a systematic way political parties’ manifestos for decades in a combined effort to create a resource for the systematic content analysis and to advance political science. The corpus is based on the work of the Manifesto Research Group (MRG) and the Comparative Manifestos (CMP) projects \parencite{budge_mapping_2001}. Classification annotations are described in the \textit{Manifesto Coding Handbook} which has evolved over the years, and provides information and instructions to the human annotators on how political parties' manifestos should be coded (latest version in \citet{volkens_manifesto_2020-1}). The handbook also includes a speficic set of policy areas or 'domains' (7) and subareas or 'subdomains' (56) which are available to annotators to use (see Figure \ref{fig1-manifestos-domains}).

For our training corpus, we use a subset of the corpus contatining 115 English Manifestos with 86,500 annotated sentences. Table \ref{table1} shows the domain codes distribution in the dataset.

\begin{table}[t]
    \centering
    \begin{tabular}{l|l}
         Domain 1 (External Relations) & 6.5\%  \\
         Domain 2 (Freedom and Democracy) & 4.42\%  \\
         Domain 3 (Political System) & 10.64\%  \\
         Domain 4 (Economy) & 25.45\%  \\
         Domain 5 (Welfare and Economy of Life) & 31.77\%  \\
         Domain 6 (Fabric of Society) & 11.20\%  \\  
         Domain 7 (Social groups) & 9.99\%  \\        
    \end{tabular}
    \caption{Domain Codes' distribution in the English subset of the Manifestos Corpus used for training the CNN classifier.}
    \label{table1}
\end{table}

\paragraph{Coronavirus (COVID-19) Press Briefings corpus}
The Coronavirus (COVID-19) Press Briefings Corpus is a collection of daily briefings on the COVID-19 status and policies from the UK and the World Health Organisation. The corpus is still in development, but we have selected example sentences from the UK and WHO which were the ones available.

During the peak of the pandemic, most countries around the world informed their citizens of the status of the pandemic (usually involving an update on the number of infection cases, number of deaths) and other policy-oriented decisions about dealing with the health crisis, such as advice about what to do to reduce the spread of the epidemic. At the moment the dataset includes briefings covering announcements between March 2020 and August 2020 from the UK (England, Scotland, Wales, Northern Ire-land) and the World Health Organisation (WHO) as follows:

\begin{itemize}
    \item UK - England: Daily Press Briefings by UK Government between 12 March 2020 – 30 Au-gust 2020 (150 briefings in total, 13,050 sentences )
    \item UK - Scotland: Daily Press Briefings by Scottish Government between 3 March 2020 - 30 August 2020 (167 briefings in total, 14,529 sentences)
    \item UK - Wales: Daily Press Briefings by Welsh Government between 23 March 2020 - 30 August 2020 (146 briefings in total, 12,702 sentences)
    \item UK - Northern Ireland: Daily Press Briefings by N. Ireland Assembly between 23 March 2020 - 30 August 2020 (130 briefings in total, 11,310 sentences)
    \item World Health Organisation: Press Briefings occurring usually every 2 days between 22 January 2020 - 30 August 2020 (124 briefings in total, 10,788 sentences)
\end{itemize}

\section{Neural Network Architecture for Topic Classification}
\paragraph{Continuous semantic text representations (Embeddings)}

We obtained pre-trained context-free word embeddings for English (Word2Vec:  \parencite{mikolov_efficient_2013}, GloVe: \parencite{pennington_glove:_2014}). Word2Vec uses a shallow neural network model to learn word associations from a large corpus of text. Once trained, such a model can detect synonymous words or suggest additional words for a partial sentence. 

\textbf{Word2Vec} uses a neural network model to learn word associations from a large corpus of text. Once trained, such a model can detect synonymous words or suggest additional words for a partial sentence. 

\textbf{GloVe} is an unsupervised learning model for obtaining vector representations for words. This is achieved by mapping words into a meaningful space where the distance between words is related to semantic similarity. Training is performed on aggregated global word-word co-occurrence statistics from a corpus, and the resulting representations showcase interesting linear substructures of the word vector space.

We also obtained word embeddings for more context-sensitive word embeddings, namely ELMo \parencite{peters_deep_2018} and BERT \parencite{devlin_bert_2019}.

\textbf{ELMo} is a deep contextualized word representation that models both (1) complex characteristics of word use (e.g., syntax and semantics), and (2) how these uses vary across linguistic contexts (i.e., to model polysemy). These word vectors are learned functions of the internal states of a deep bidirectional language model (biLM), which is pre-trained on a large text corpus. They can be easily added to existing models and significantly improve the state of the art across a broad range of challenging NLP problems, including question answering, textual entailment and sentiment analysis. 

\textbf{BERT} is  a deeply bidirectional, unsupervised language representation, pre-trained using only a plain text corpus. It  includes a variant that uses the English Wikipedia with 2.5 million words. Unlike previous context-free models, which generate a single word embedding representation for each word in the vocabulary, BERT takes into account the context for each occurrence of a given word, providing a contextualised embedding that is different for each sentence.

\paragraph{Convolutional Neural Networks}

Since \citet{kim_convolutional_2014}'s paper outlining the idea of using CNNs for text classification (traditionally used for recognising visual patterns from images), CNNs have achieved very good performance in several text classification tasks \parencite{poria_deep_2015, bilbao-jayo_political_2018}. CNNs involve convolutional operations of moving frames or windows (filter sizes) which analyse and reduce different overlapping regions in a matrix, to extract different features. The ability to also bootstrap word embeddings in this type of neural network make it an excellent candidate for extracting knowledge and classifying non-annotated texts.

We therefore set up 4 variations of the CNN classifier M1, M2, M3, M4 as follows:

\begin{enumerate}
      \item Word vectors of the training dataset sentences are created using one of the following word embeddings: Word2Vec (M1), GloVe (M2), ELMo (M3) and BERT (M4). Sentences are fed as sequences of words, then mapped to indexes, then a sequence of word vectors. We have chosen 300 as the word vector size and 60 x d for the space where the convolution operations can be performed.
    \item Vectors are fed to the neural network (CNN). we then perform convolution operations with 100 filters and three different filter sizes (2 x d, 3 x d, and 4 x d). We reduce the dimensionality of the feature maps generated by each group of filters using \textit{1-max-pooling}, which are consequently concatenated  \parencite{boureau_theoretical_2010}.
    \item A dropout rate of 0.5 is applied \parencite{srivastava_dropout_2014} as regularisation to prevent overfitting.
    \item The layer with \textit{softmax} computes the probability distribution over the labels.
    \item We perform optimization using the Adam optimiser with the parameters of the original manuscript \parencite{kingma_adam_2017}.
    
\end{enumerate}

Note that this is a sentence-level topic classifier basing its predictions by taking into account only the information local within the sentence.

\section{Evaluation}

For our training corpus, we use a subset of the corpus containing 115 English Manifestos with 86,500 annotated sentences. Table \ref{table1} shows the domain codes distribution in the dataset. In order to evaluate the different architectures, we divided our training dataset in 2 different subsets: training and validation sets (85\%) and test set (15\%). Typically, we have used a validation set (or development test set) separate from the test set, to ensure correct evaluation and that our model(s) do not overfit, thus ensuring how each domain is classified and that the evaluation is robust.

We performed 4 experiments, one for each combination of CNN and word embeddings:

\begin{itemize}
    \item M1: CNN with Word2Vec
    \item M2: CNN with GloVe
    \item M3: CNN with ELMo
    \item M4: CNN with BERT
\end{itemize}

As shown in Table~\ref{table3}, the performance of the classifier improves when more context-sensitive word embeddings are used. Using BERT with CNN (M4) seems to provide a substantial increase in accuracy and F1, whereas using ELMo performs very well as well.

\begin{table}[t]
    \centering
    \begin{tabular}{l|l|l}
     \textbf{Experiment}   & \textbf{Accuracy} & \textbf{F1} \\
      M1  & 65.79\%  & 61.11  \\
      M2  & 68.15\%  & 64.93 \\
      M3  & 72.84\%  & 68.42  \\
      M4  & 87.52\%  & 74.68  \\
    \end{tabular}
    \caption{Domain results of all models using political manifestos}
    \label{table3}
\end{table}

\paragraph{Applying the models on the COVID-19 corpus}

We also tested the performance of the same different pre-trained models on the COVID-19 corpus.  We asked two political science scholars to annotate a subset of 20 press briefings (4 of each set), using the 7 domains of the Manifestos Project. This resulting in a dataset of 1740 manually annotated sentences, with domain distrubution as in Table \ref{table4}. Note that the pre-trained models have been trained using the annotated manifestos from the Manifestos Project, without any additional training on the press briefings corpus sentences.

\begin{table}[t]
    \centering
    \begin{tabular}{l|l}
         Domain 1 (External Relations) & 0.74\%  \\
         Domain 2 (Freedom and Democracy) & 0.47\%  \\
         Domain 3 (Political System) & 11.58\%  \\
         Domain 4 (Economy) & 33.99\%  \\
         Domain 5 (Welfare and Economy of Life) & 34.62\%  \\
         Domain 6 (Fabric of Society) & 15.02\%  \\  
         Domain 7 (Social groups) & 3.58\%  \\        
    \end{tabular}
    \caption{Manifest Project Domain Codes' distribution in the manually annotated subset of the COVID-19 corpus.}
    \label{table4}
\end{table}

As shown in Table \ref{table5}, the performance of the classifier improves when more context-sensitive word embeddings are used in the context of the COVID-19 press briefings corpus as well. Using BERT with CNN (M4) seems to provide a substantial increase in accuracy and F1, whereas using ELMo performs very well as well. As expected there is some loss of accuracy, as we are porting the classifier to a slightly different domain of political text (from manifestos to press briefings).

\begin{table}[t]
    \centering
    \begin{tabular}{l|l|l}
     \textbf{Experiment}   & \textbf{Accuracy} & \textbf{F1} \\
      M1  & 50.65\%  & 48.62  \\
      M2  & 54.18\%  & 48.82 \\
      M3  & 60.74\%  & 57.07  \\
      M4  & 68.65\%  & 64.58  \\
    \end{tabular}
    \caption{Domain results of all models using COVID-19 Press briefings corpus}
    \label{table5}
\end{table}

\section{Conclusion}
In this paper, we built a sentence-level political discourse classifier using existing human expert annotated corpora of English political manifestos from the Manifestos Project \parencite{volkens_manifesto_2020}. We tested the accuracy and performance of a neural networks classifier (CNN) using different word embeddings as part of the word to vector mapping and we showed that sentence-level CNN classifiers combined with transformers like BERT outperform models with other embeddings (Word2Vec, Glove, ELMo). We then applied the same pre-trained models to a different set of text, the COVID-19 Press Briefings Corpus. We observe similar patterns in the accuracy and F1 scores, and additionally show that it is possible to use a pre-trained classifier to conduct automatic classification on different political texts without additional training

In the future, we aim to conduct similar experiments also considering the 'subdomain' categories of the Manifesto Corpus Annotations. We also look forward to re-running these experiments for other languages in the Manifestos project, testing the language-agnostic advantage of word embeddings and see if we could obtain different results.

\section{Ethics statement}
This paper follows the AAAI Publications Ethics and Malpractice Statement and the AAAI Code of Professional Conduct. We use publicly available text data to ensure transparency and reproducibility of the research. Additionally, all code will be available as open source code (on github.com) at the end of the submission and reviewing process.

The paper suggests ways to automatically extract topic information from political discourse texts, employing deep learning methods which are usually associated with artificial intelligence and ethical considerations around them. We do not envisage any ethical, social and legal considerations arising from the work outlined in this study, such as impact of AI on humans, on economic growth, on inequality, amplifying bias or undermining political stability or other issues described in recent reports on ethics in AI (see for example \parencite{bird_ethics_2020}).


\section*{Acknowledgements}
The author would like to acknowledge the support of the Business and Local Government Data Research Centre (ES/S007156/1) funded by the Economic and Social Research Council (ESRC) whilst undertaking this work.


\renewcommand{\refname}{\spacedlowsmallcaps{References}} 

\printbibliography

\newpage

\tableofcontents 


\listoffigures 

\listoftables 

\end{document}